\documentclass[10pt,twocolumn,letterpaper]{article}

\usepackage[pagenumbers]{cvpr} % To force page numbers, e.g. for an arXiv version

\usepackage{graphicx}
\usepackage{amsmath}
\usepackage{amssymb}
\usepackage{booktabs}

\usepackage[pagebackref,breaklinks,colorlinks]{hyperref}

\usepackage[capitalize]{cleveref}
\crefname{section}{Sec.}{Secs.}
\Crefname{section}{Section}{Sections}
\Crefname{table}{Table}{Tables}
\crefname{table}{Tab.}{Tabs.}

\usepackage{graphicx}
\usepackage{multirow}
\usepackage{bbding}
\usepackage{bbm}
\usepackage{algorithm}
\usepackage{algorithmic}
\usepackage{makecell}
\makeatletter
\renewcommand{\maketag@@@}[1]{\hbox{\m@th\normalsize\normalfont#1}}%
\makeatother

\DeclareMathOperator*{\argmax}{arg\,max}

\DeclareMathOperator*{\argsort}{arg\,sort}

\newcommand{\red}[1]{\textcolor{red}{#1}}
\newcommand{\blue}[1]{\textcolor{blue}{#1}}  % blue标签有专用，请勿更换

\def\cvprPaperID{264} % *** Enter the CVPR Paper ID here
\def\confName{CVPR}
\def\confYear{2023}

\begin{document}

% \title{Contrastive Semi-Supervised Learning with Complementary Labeling}
\title{Boosting Semi-Supervised Learning with Contrastive Complementary Labeling}
\author{Qinyi Deng\footnotemark[1], Yong Guo\footnotemark[1], Zhibang Yang, Haolin Pan, Jian Chen \\
South China University of Technology\\
%Institution1 address\\
{\tt\small sedengqy@mail.scut.edu.cn,
guoyongcs@gmail.com,}\\
{\tt\small \{se202030482287, sephl\}@mail.scut.edu.cn, ellachen@scut.edu.cn}
}
\maketitle

\renewcommand{\thefootnote}{\fnsymbol{footnote}}
\footnotetext[1]{Authors contributed equally.}
%\footnotetext[2]{Corresponding author.}

\begin{abstract}
Semi-supervised learning (SSL) has achieved great success in leveraging a large amount of unlabeled data to learn a promising classifier.
A popular approach is pseudo-labeling that generates pseudo labels only for those unlabeled data with high-confidence predictions.
As for the low-confidence ones, existing methods often simply discard them because these unreliable pseudo labels may mislead the model.
Nevertheless, we highlight that these data with low-confidence pseudo labels can be still beneficial to the training process. Specifically, although the class with the highest probability in the prediction is unreliable, we can assume that this sample is very unlikely to belong to the classes with the lowest probabilities.
In this way, these data can be also very informative if we can effectively exploit these complementary labels, i.e., the classes that a sample does not belong to.
Inspired by this, we propose a novel Contrastive Complementary Labeling (CCL) method that constructs a large number of reliable negative pairs based on the complementary labels and adopts contrastive learning to make use of all the unlabeled data.
Extensive experiments demonstrate that CCL significantly improves the performance on top of existing methods. More critically, our CCL is particularly effective under the label-scarce settings. For example, we yield an improvement of 2.43\% over FixMatch on CIFAR-10 only with 40 labeled data.
\end{abstract}

\section{Introduction}

Deep networks have been the workhorse of many computer vision tasks, including image classification~\cite{he2016deep,guo2021towards,simonyan2014very,guo2022towards,guo2020breaking,chen2022automatic,guo2019nat,chen2021contrastive}, semantic segmentation~\cite{long2015fully,liu2020dynamic}, and many other areas~\cite{xu2022downscaled,guo2020closed,niu2021adaxpert,guo2022pareto,guo2020hierarchical,guo2019auto,cao2018adversarial,liu2021deep}.
Recently, semi-supervised learning (SSL) has achieved great success in training deep models on large-scale datasets without expensive labeling costs~{\cite{sohn2020fixmatch,zhang2021flexmatch, yang2022class, duan2022mutexmatch}}.
Compared to the standard supervised learning scheme, SSL can unleash the power of large amounts of unlabeled data to learn better classifiers, which is especially effective under a labels-scarce setting.
One technique that is widely used in SSL methods is pseudo-labeling, which generates pseudo labels for unlabeled data to provide additional supervision information.
In practice, the quality of pseudo labels is critical to the performance of SSL methods.
To avoid low-quality pseudo labels hindering training, a common strategy is to generate pseudo labels only for unlabeled data with high-confidence predictions, while discarding other unlabeled data entirely.

However, this simple selective strategy leads to limited utilization of unlabeled data, as these discarded unlabeled data may contain information that can be further exploited.
Some advanced methods ~\cite{li2021comatch, yang2022class, kim2021selfmatch} try to use contrastive learning to further mine the information in low-confidence unlabeled data, to effectively use unlabeled data.
One representative work, CCSSL~\cite{yang2022class}, directly uses low-confidence unlabeled data to construct pairs and further computes supervised contrastive loss~\cite{khosla2020supervised}.
As a result, CCSSL improves SSL methods by mining extra information for low-confidence unlabeled data to a certain extent, but still inevitably leads to suboptimal models.
Since CCSSL may mistakenly push data belonging to the same class away based on the low-confidence pseudo labels, the discriminative power of the model will be weakened.

To address the issue, we hope to reduce possible noises when using low-confidence unlabeled data for training.
Intuitively, we should keep data that may belong to the same class close to each other in the feature space.
However, unlabeled data with low-confidence often produce a very noisy signal about which classes they should belong to, i.e., the classes with the highest probability in predictions are considered unreliable.
On the contrary, we can always confidently keep the data with low-confidence away from the region of the class with the lowest probability. 
For each low-confidence data, the class it is most unlikely to belong to is considered as its complementary label.
From this point of view, if we can better make use of these complementary labels, the low-confidence unlabeled data can help to learn a discriminative SSL model.

Inspired by this, we seek to leverage the complementary labels of data with low-confidence. 
Nevertheless, these complementary labels cannot be directly used to compute the cross-entropy loss. 
Instead, we propose a Contrastive Complementary Labeling~(CCL) method which constructs reliable negative pairs for low-confidence data based on their complementary labels and further performs contrastive learning.
Specifically, we construct reliable negative pairs between either two high-confidence samples or one high-confidence sample together with a low-confidence sample. We highlight that these pairs are reliable because both the labels of high-confidence samples (the former case) and the complementary labels of low-confidence samples (the latter case) are considered reliable.
But unlike standard complementary labels that only take the class with the lowest probability, we select several low-probability classes as complementary labels for low-confidence unlabeled data.
This pair construction strategy enables CCL to construct a large number of reliable negative pairs, which are beneficial for the subsequent contrastive learning.
As for positive pairs, we only consider the samples that belong to the same class or are generated through different augmentations from the same image.
In this way, both the negative pairs and positive pairs become reliable. More critically, unlike most existing methods that directly discard those low-confidence samples, we highlight that our CCL method takes all the unlabeled data into account.
Extensive experiments show that CCL can be easily integrated with existing advanced SSL methods and significantly outperform the baselines under the label-scarce settings.

Our contributions can be summarized as follows:
\vspace{-3 pt}
\begin{itemize}
\setlength\itemsep{0 em}
\item We propose a novel Contrastive Complementary Labeling~(CCL) method that constructs reliable negative pairs based on the complementary labels, i.e., the classes that a sample does not belong to. Indeed, our CCL effectively exploits low-confidence samples to provide additional information for the training process.
\item 
We develop a complementary labeling strategy to construct reliable negative pairs. Specifically, for low-confidence data, we first select multiple classes with the lowest probability as complementary labels. Then reliable complementary labels are applied to construct a large number of negative pairs, which greatly benefits the contrastive learning.
% thereby achieving a promising trade-off between the number and correctness of negative pairs.
\item Extensive experiments on multiple benchmark datasets show that CCL can be easily integrated with existing advanced pseudo-label-based end-to-end SSL methods (e.g., FixMatch) and finally yield better performance. Besides, under the label-scarce settings, CCL effectively unleashes the power of data with low-confidence, thus achieving a significant improvement of 2.43\%, 6.35\%, and 1.78\% on CIFAR-10, STL-10, and SVHN only with 40 labeled data, respectively. 
\end{itemize}

\begin{figure*}[ht]
\centering
\includegraphics[width=1\linewidth]{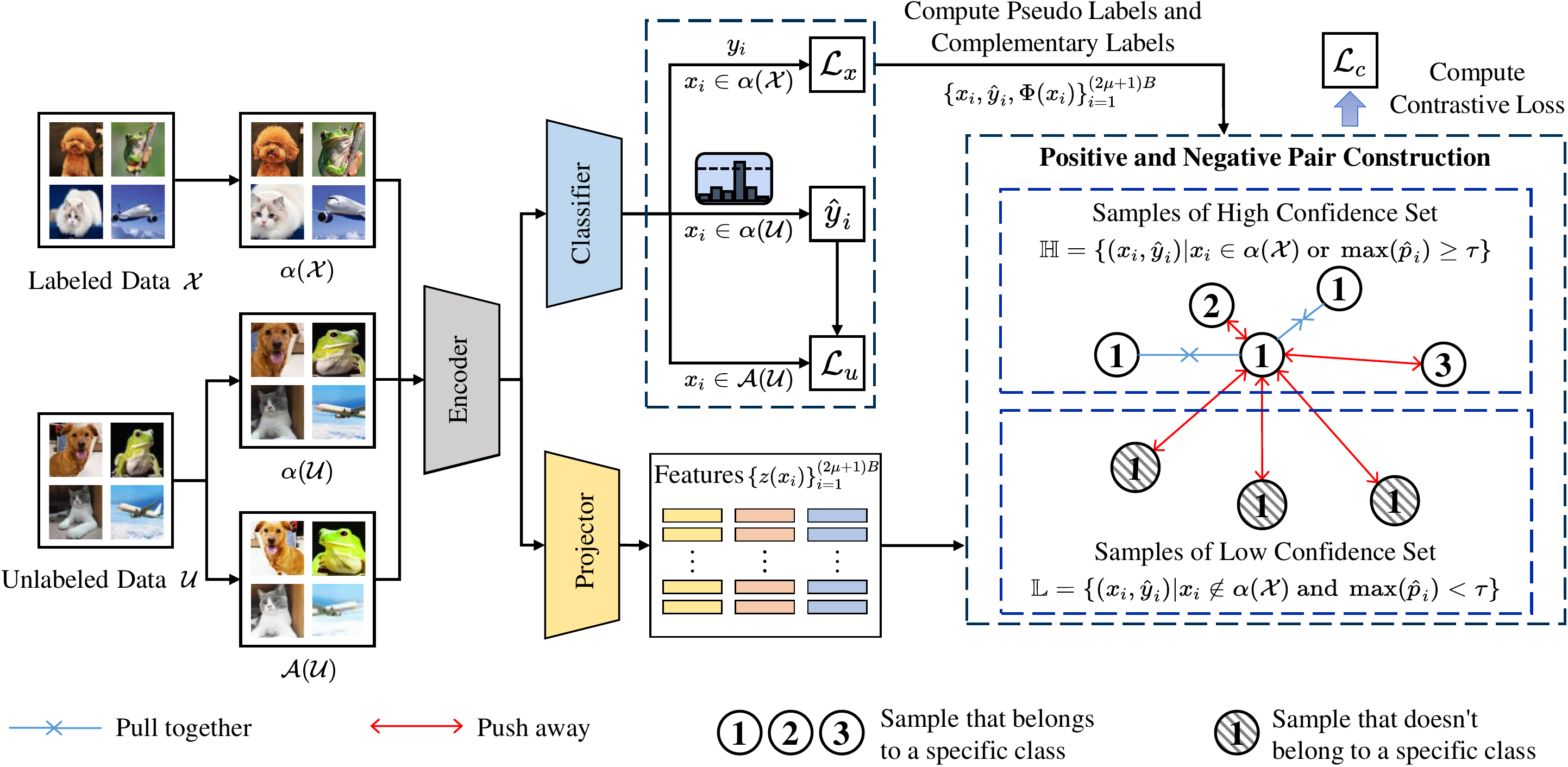} %Structure.jpg
\caption{The schema of the proposed Contrastive Complementary Labeling~(CCL) method. Given a batch of labeled data $\mathcal{X}$ and unlabeled data $\mathcal{U}$, the input data $x_i$ will go through the classifier to generate pseudo labels $\hat y_i$ and obtain complementary labels via $\Phi(x_i)$. Then we divide the input data with their pseudo labels into the high-confidence set $\mathbb{H}$ and the low-confidence set $\mathbb{L}$. The projector head helps map features to the normalized embeddings to compute the contrastive loss $\mathcal{L}_c$. By constructing positive pairs and negative pairs with complementary labels, we focus on leveraging reliable information from those low-confidence unlabeled data.}
\label{figure: structure}
\end{figure*}

\section{Related Work}
\subsection{Semi-Supervised Learning}
Semi-supervised learning (SSL) mainly include consistency regularization method and pseudo-labeling methods, which aims at taking advantage of unlabeled data to improve performance.
Specifically, consistency regularization methods \cite{jeong2019consistency,  rasmus2015semi, laine2016temporal, tarvainen2017mean, xie2020unsupervised,liu2021conditional, ke2019dual,zhuang2018discrimination,liu2021discrimination, park2018adversarial,guo2021content, athiwaratkun2018there, zhang2020wcp, kuo2020featmatch,guo2022improving, liu2020unbiased,guo2018double,guo2020multi,niu2021disturbance,cao2020improving,han2016online} construct different views from the same image, based on the assumption that perturbation of images can not cause significant change of the classifier output. 
Pseudo-labeling methods~\cite{nassar2021all, pham2021meta, zheng2022simmatch, zhao2022lassl, zhou2021instant, tang2021humble, xu2021end} rely heavily on confidence threshold, where choosing the appropriate threshold is important to achieving high performance.

Recently, ReMixMatch \cite{berthelot2019remixmatch} uses weak augmentation to create pseudo labels and enforce consistency against strong augmentation samples. FixMatch \cite{sohn2020fixmatch} and UDA \cite{xie2020unsupervised} rely on a fixed threshold during training, and only retain the unlabeled data whose prediction probability is above the threshold. Based on the limitations of the threshold setting of FixMatch, FlexMatch \cite{zhang2021flexmatch} proposes Curriculum Pseudo Labeling to adjust thresholds flexibly for different classes, which effectively leverages unlabeled data. With the help of contrastive learning in SSL, CoMatch~\cite{li2021comatch} and CCSSL~\cite{yang2022class} cluster samples with similar pseudo labels to alleviate the problem that the model cannot mine information from unreliable low-confidence unlabeled data. 
Inspire by these works, CCL further uses multiple complementary labels to construct reliable negative pairs, and then provides additional discriminative information for contrastive learning, thereby facilitating the utilization of unlabeled data.

\subsection{Contrastive Learning}
Contrastive learning methods have been fully developed in recent years. \cite{hinton2002training, hyvarinen2005estimation, gutmann2010noise} propose and refine Noise Contrastive Estimation~(NCE).  
\cite{oord2018representation} proposes InfoNCE to help the model capture valuable information for future prediction. In self-supervised learning, contrastive learning is widely applied and has shown strong performance \cite{ chen2020simple, chen2020big, he2020momentum, grill2020bootstrap, chen2021exploring, zbontar2021barlow, chen2021empirical, caron2021emerging }.~%是不是太多了，太长了
Self-supervised contrastive learning constructs a latent feature space for the downstream task. To leverage label information effectively, SupCon \cite{khosla2020supervised} clusters the same class samples and pulls apart different classes. Combining SupCon, full-supervised learning obtains a significant improvement in image classification. \par
Recently, some methods have been successfully applied in different domains by combining contrastive learning. For example, \cite{alonso2021semi} proposes a pixel-level contrastive learning scheme for SSL semantic segmentation with a memory bank for pixel-level features from labeled data. \cite{pan2022improving} improves fine-tuning of self-supervised pretrained models with contrastive learning. In the SSL classification task, \cite{zhang2022semi} and \cite{kim2021selfmatch} combine self-supervised contrastive learning with cross-entropy and consistency regularization to achieve better performance, respectively. 
Our method differs from them by constructing reliable negative pairs based on complementary labels in a novel strategy, which is beneficial for subsequent contrastive learning.

\subsection{Complementary Labeling}

Complementary labels indicate which classes the image does not belong to. Compared with ordinary labels, complementary labels are sometimes easily obtainable, especially when the class set is relatively large. For classification with complementary labels, \cite{ ishida2017learning } provides an unbiased estimation for the expected risk of classification with true labels and proposes a framework to minimize it. \cite{ishida2019complementary} modifies the theories and frameworks in \cite{ishida2017learning} and improves its performance. \cite{Yu_2018_ECCV} further addresses the problem of learning with biased complementary labels. It is worth mentioning that \cite{ishida2017learning, ishida2019complementary} only consider using a single class with the lowest probability as a complementary label. Different from them, CCL considers multiple unlikely classes as complementary labels for each low-confidence data to provide more discriminative information.

However, complementary labels can not be directly used to compute cross-entropy. MutexMatch~\cite{duan2022mutexmatch} proposes to train a distinct classifier TNC for predicting complementary labels, thus helping to learn the representation of unlabeled data from a mutex perspective and greatly boosting SSL methods.
Instead of incorporating complementary labels into computing cross-entropy loss, CCL naturally constructs reliable negative pairs based on complementary labels and further computes the contrastive loss.

\section{Proposed Method}
In this paper, we seek to improve SSL by leveraging reliable information from those low-confidence unlabeled data. To this end, we propose a Contrastive Complementary Labeling (CCL) method that generates a large number of reliable negative pairs based on the classes that a sample does not belong to, namely complementary labels. For clarity, in \Cref{preliminary}, we first discuss some important notations and preliminaries. Then, in \Cref{Main method}, we develop the rules on how to construct reliable negative pairs and incorporate them into the contrastive learning scheme. The overview of the proposed CCL method is shown in \Cref{figure: structure}.

\subsection{Notations and Preliminaries}
\label{preliminary}
We consider the semi-supervised setting where
the training data consists of some labeled data along with a large number of unlabeled data. Let $\mathcal{X}{=}\left\{\left(x_{i}, y_{i} \right)\right\}_{i=1}^{B}$ be a batch of $B$ labeled samples where $x_i$ and $y_i$ are the training sample and the corresponding label, respectively. Let $\mathcal{U}{=}\left\{u_{i}\right\}_{i=1}^{\mu B}$ be a batch of $\mu B$ unlabeled samples where $\mu$ denotes the relative sizes of $\mathcal{X}$ and $\mathcal{U}$. Following~\cite{sohn2020fixmatch}, we use the weak augmentation $\alpha(\cdot)$ to transform all the training data and additionally use a strong augmentation $\mathcal{A}(\cdot)$ to produce an extra view of the unlabeled data. In total, we consider three kinds of data, i.e., $\alpha(\mathcal{X})$, $\alpha(\mathcal{U})$, and $\mathcal{A}(\mathcal{U})$.

In this paper, we consider a deep network that consists of an encoder, a classifier head, and a projector head.
The encoder extracts features of images and the classifier further produces the predicted logits.
We use a projector head to map features to the normalized embeddings to compute the contrastive loss. 
When generating pseudo labels, we first obtain the prediction $f(\cdot)$ of the weakly augmented view and enforce the corresponding strongly augmented view to share the same logits and pseudo label.
To achieve this, we define a function $\psi(x)$ that returns the weakly augmented view of any given data $x$.
Formally, the predictions of different data can be formulated by:
\begin{equation}
\hat p_{i}=\left\{ \begin{array}{*{35}{l}}
    f(x_i)  & \text{if }{{x}_{i}} \in \alpha(\mathcal{X}),  \\
    f(\psi(x_i)) & \text{otherwise}. \\
    % -1 & \text{otherwise} \\
\end{array} \right. 
\label{equation of predicted class distribution}
\end{equation}
% where $\psi(x)$ returns the weakly augmented view of $x$. 
% $f(x)$ obtains the model's prediction of $x$. 

Based on \cref{equation of predicted class distribution}, we further compute the (pseudo) label for each sample. 
For labeled data, we directly take the ground-truth labels $y_i$.
For unlabeled data, we focus on the largest item of $\hat p_i$ and use a confidence threshold $\tau$ to select those high-confidence samples with $\max ({{\hat p}_{i}})\ge \tau$. In practice, we leave the other low-confidence samples with $\hat y_i=-1$ since their pseudo labels are often unreliable.
In summary, the labels $\hat y_{i}$ of different data becomes:
\begin{equation}
\hat y_{i}=\left\{ \begin{array}{*{35}{l}}
    y_i  & \text{if }{{x}_{i}} \in \alpha({\mathcal{X}}),  \\
    \argmax(\hat {p}_i) & \text{else if}~ \max ({{\hat p}_{i}})\ge \tau,   \\
    -1 & \text{otherwise}. \\
\end{array} \right. 
\label{equation of pseudo label}
\end{equation}
% where $y_i$ is ground-truth labels for labeled data. $\tau$ denotes the threshold for determining the prediction confidence.

In summary, all the labeled data and unlabeled data, along with their labels, can be written as $\mathbb{Q}=\left\{\left(x_{i}, \hat{y}_{i}\right) \right\}_{i=1}^{(2 \mu+1) B}$,
where $x_{i} \in \alpha(\mathcal{X}) \cup \alpha(\mathcal{U}) \cup \mathcal{A}(\mathcal{U})$.
Based on the above definitions, we take both labeled data and the high-confidence unlabeled data as the high-confidence set: 
\begin{equation}
     \mathbb{H}=\{\left(x_{i}, \hat{y}_{i}\right) | x_i \in \alpha(\mathcal{X})  {~\rm or~} \max ({{\hat p}_{i}})\ge \tau \}.
\label{equation of high-confidence unlabeled data}
\end{equation}
And leave the rest as the low-confidence set $\mathbb{L}=\mathbb{Q} \setminus \mathbb{H}$. In practice, most SSL methods simply discard the use of noisy $\mathbb{L}$, however, these low-confidence data may still provide some useful information.

\subsection{Contrastive Complementary Labeling}
\label{Main method}
Besides the samples in high-confidence set $\mathbb{H}$, we seek to further exploit those low-confidence samples in $\mathbb{L}$ to aid the training.
Although taking the class with the largest probability as the label is very noisy for the low-confidence samples, those classes with particularly low probabilities should still be relatively reliable to indicate which classes a sample does not belong to. 
For clarity, we refer to these unlikely classes as \emph{complementary labels} in this paper.

Although these complementary labels become potentially informative, it is non-trivial to directly use them to compute the cross-entropy loss. 
% Instead, we can easily tell whether a low-confidence sample does not belong to the same classes with a larger number of  high-confidence samples in $\mathbb{H}$.
%Instead, we can easily construct negative pairs between a low-confidence sample and a larger number of high-confidence samples in $\mathbb{H}$ that belong to the unlikely classes.
Instead, we can easily construct negative pairs between a low-confidence sample and a large number of high-confidence samples in $\mathbb{H}$ that belong to classes in the complementary label of the low-confidence sample.
% defined by the complementary labels.
Interestingly, these negative pairs can be naturally incorporated into the contrastive learning scheme~\cite{khosla2020supervised}. Inspired by this, we propose a Contrastive Complementary Labeling (CCL) method that constructs reliable negative pairs based on the complementary labels. 
The detailed training method is shown in \Cref{algorithm: CSSC}.
In the following, we will discuss the construction of complementary labels and the specific strategy to construct reliable negative and positive pairs upon them. 

% \deng{In most existing SSL methods, the model pays no attention to low-confidence data, which is more likely to be noise. However, by setting a high confidence threshold to divide samples, a lot of information is often discarded in the early stage of training. Meanwhile, an unreliable prediction in classification is usually confused among only a few classes instead of all classes. Therefore, we adopt contrastive learning with complementary labels in constructing pairs for relatively reliable low-confidence ones.}
% \yong{discuss importance of low-confidence samples here!}

% It is obvious that reliable pseudo labels can provide helpful information for better discrimination. However, directly leveraging all unlabeled data is unreliable, as the model's predictions may contain a lot of noise. Therefore, to make better use of data with different confidence levels, we first pass the unlabeled data through the confidence threshold $\tau$ for separation according to the model’s prediction confidence.

\begin{algorithm}[!t]
\caption{Training method of \textbf{Contrastive Complementary Labeling (CCL)}. We obtain multiple complementary labels for each sample and build reliable negative pairs based on them to compute the contrastive loss.}
\label{algorithm: CSSC}
\textbf{Input}: Labeled data ${\mathcal{X}}$, unlabeled data ${\mathcal{U}}$, the function to obtain complementary labels for a sample $\Phi(\cdot)$, the function to return the weakly augmented view of a sample $\psi(\cdot)$, the number of classes in complementary labels ${k}$. \\
\vspace{-13 pt}
\begin{algorithmic}[1]
\STATE \emph{// Obtain pseudo labels for unlabeled data}
\FOR{$ x_i \in \mathcal{U} $}
    \STATE Obtain predicted class distribution $\hat p_{i}$ via \cref{equation of predicted class distribution}
    \STATE Compute the pseudo label $\hat{y}_{i}$ using \cref{equation of pseudo label}
\ENDFOR
\STATE \emph{// Construct negative pairs and positive pairs}
\STATE Divide $\mathbb{Q}$ into the high-confidence set $\mathbb{H}$ and the low-confidence set $\mathbb{L}$ via \cref{equation of high-confidence unlabeled data}
\STATE Obtain complementary labels via \\
$\Phi(x_i)=\left\{ \begin{array}{*{35}{l}}
    \{ c \in \mathcal{O}_{i} | c \neq \hat{y}_i \} & \text{if }{{x}_{i}} \in \mathbb{H},  \\
    \{c \in \mathcal{O}_{i} | \mathcal{O}_{i}(c) < k \} & \text{otherwise}.  \\
\end{array} \right.$
\STATE Construct negative pairs using \\
$\mathbb{N}({{x}_{i}})=\{{{x}_{j}} \in \mathbb{Q} |{\hat y_{j}}\in {\Phi(x_i)} {~\rm or~} \hat y_i \in {\Phi(x_j)} \}$.
\STATE Construct positive pairs using \\
{$\mathbb{P}(x_i){=}\left\{ \begin{array}{*{35}{l}}
\{{{x}_{j}}\in \mathbb{Q}{\setminus}\{x_i\}|{\hat{y}_{i}}{=}{\hat{y}_{j}}\} & \text{if}\text{ } x_i\in \mathbb{H},  \\
\{{{x}_{j}}\in \mathbb{Q}{\setminus}\{x_i\}| \psi(x_i){=}\psi(x_j) \} & \text{otherwise}. \\
\end{array} \right.$}
\STATE \normalsize{Compute contrastive loss $\mathcal{L}_c$ via \cref{equation of contrastive loss}}
\STATE \normalsize{Compute total loss $\mathcal{L}=\mathcal{L}_{x}+ \mathcal{L}_{u}+\mathcal{L}_{c}$ to update model}
\end{algorithmic}
\end{algorithm}

\paragraph{Constructing negative pairs with complementary labels.}
% Although the model may not accurately predict what class a low-confidence unlabeled data belongs to, it is easier to guess what class it does not belong to.

For any sample $x_i \in \mathbb{Q}$, we seek to construct the complementary labels based on its predicted class distribution $\hat p_i$. Since we seek to identify the complementary classes that a sample does not belong to, we sort $\hat p_i$ in the ascending order and obtain the indices of classes via ${{\mathcal{O}}_{i}=\argsort(\hat{p}_i)}$. 
% \yong{Discuss why consider multiple unlikely classes. Highlight that it is different from existing work.}
Actually, the standard complementary labeling~\cite{ishida2017learning} only selects the first one, i.e., the class with the lowest probability, as the complementary label. Nevertheless, given a number of classes in total, e.g., $C{=}10$ for CIFAR-10, the second or even the first five items in $\mathcal{O}$ may also have very low probabilities and can be potentially used to construct the complementary labels.
% resulting in a limited number of negative pairs, which hinders subsequent contrastive learning.
Inspired by this, we propose a flexible complementary labeling strategy that selects multiple unlikely classes as complementary labels for each low-confidence sample.
As for the high-confidence sample $x \in \mathbb{H}$, we directly take all the classes other than its ground-truth label as the complementary labels.
For convenience, we use ${\mathcal{O}}_{i}(c)$ to denote the position of class $c$ in $\mathcal{O}_{i}$. Formally, the complementary label of $x_i$ can be written by:
\begin{equation}
\Phi(x_i)=\left\{ \begin{array}{*{35}{l}}
    \{ c \in \mathcal{O}_{i} | c \neq \hat{y}_i \} & \text{if }{{x}_{i}} \in \mathbb{H},  \\
    \{c \in \mathcal{O}_{i} | \mathcal{O}_{i}(c) < k \} & \text{otherwise},  \\
\end{array} \right.  
\label{equation of complementary labels}
\end{equation}
where $k$ denotes the number of unlikely classes selected as complementary labels (see ablations on $k$ in \Cref{figure: different k}).

\begin{figure}[t]
  \centering
  \begin{subfigure}{0.49\linewidth}
  \centering
  \hspace{-4.5mm}
    \includegraphics{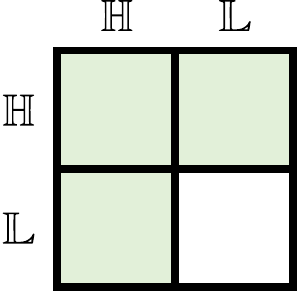}
    \caption{The data used to construct negative pairs in the proposed CCL.}
    \label{subfig:Selection of negative Pairs}
  \end{subfigure}
  \hfill
  \begin{subfigure}{0.49\linewidth}
  \centering
  \hspace{-4.5mm}
    \includegraphics{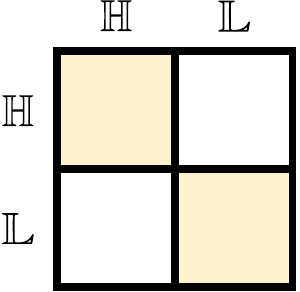}
    \caption{The data used to construct positive pairs in the proposed CCL.}
    \label{subfig:Selection of positive Pairs}
  \end{subfigure}
  \caption{The data usage matrix for constructing pairs. The shaded regions in (a) represent three cases of constructing negative pairs by \cref{equation of negative pairs}, and the shaded regions in (b) represent two cases of constructing positive sample pairs by \cref{equation of positive pairs}. Adopting this data usage strategy can make better use of low-confidence data.}
  \label{fig:Selection of Pairs}
\end{figure}

% To naturally and effectively incorporate the obtained complementary labels into contrastive learning, we adopt a novel strategy to construct negative pairs, which contains three different cases as shown in \Cref{subfig:Selection of negative Pairs}.
Based on the complementary labels, we further discuss how to construct reliable negative pairs. As shown in \Cref{subfig:Selection of negative Pairs}, we construct reliable negative pairs between either two high-confidence samples (i.e., left top corner) or one high-confidence together with a low-confidence sample (i.e., left bottom and right top corner). 
In the former case, for any two samples $x_i, x_j \in \mathbb{H}$, they become a negative pair as long as one of them belongs to the complementary label of another, i.e., ${\hat y_{j}}\in {\Phi(x_i)} {~\rm or~} \hat y_i \in {\Phi(x_j)}$.
As for the latter case, for any low-confidence sample $x \in \mathbb{L}$, we select all the samples in $\mathbb{H}$ that belong to its complementary label set $\Phi(x)$ to construct negative pairs. We highlight that these pairs are reliable because both the labels of high-confidence samples and the complementary labels of the considered low-confidence sample are reliable. Since the data usage matrix in \Cref{subfig:Selection of negative Pairs} is symmetric, we can always construct a negative pair as long as one sample belongs to the complementary labels of another one, which shares the same rule as the case for two high-confidence samples. Formally, the set of negative pairs for any sample $x \in \mathbb{Q}$ becomes:
\begin{equation}
    \mathbb{N}({{x}_{i}})=\{{{x}_{j}} \in \mathbb{Q} |{\hat y_{j}}\in {\Phi(x_i)} {~\rm or~} \hat y_i \in {\Phi(x_j)} \}.
\label{equation of negative pairs}
\end{equation}

As mentioned above, for any low-confidence sample, we can always go through the whole high-confidence set $\mathbb{H}$ to construct negative pairs. In this sense, the proposed CCL method is able to construct a large number of reliable negative pairs and thus provide richer information for contrastive learning. From this point of view, we highlight that the improvement achieved by our CCL mainly stems from a large amount of reliable negative pairs.

\paragraph{Constructing positive pairs.}
% To avoid unreliable unlabeled data affecting the effect of contrastive learning, inspired by \cite{khosla2020supervised} and \cite{oord2018representation}, we only construct positive pairs for two relatively reliable cases.
Inspired by \cite{khosla2020supervised} and \cite{oord2018representation}, we only construct positive pairs for two reliable cases to reduce possible noise from the unreliable predictions of unlabeled data as shown in \Cref{subfig:Selection of positive Pairs}.
First, under the guidance of $\hat y$, any two samples $x_i, x_j \in \mathbb{H}$ from the same class, i.e., $\hat y_i=\hat y_j$, are considered as a positive pair and should be pulled closer to each other.
Second, for the low-confidence set $\mathbb{L}$, we construct positive pairs for samples $x_i$ and $x_j$ only when they are two augmented views of the same image. 
Since we use $\psi(\cdot)$ to retrieve the weakly augmented view of an image, the above condition can be equivalently cast into sharing the same weakly augmented view, i.e., $\psi(x_i) = \psi(x_j)$. For all the other cases with one high-confidence sample and one low-confidence sample (blank regions in~\Cref{subfig:Selection of positive Pairs}), we do not construct any positive pairs since they are always unreliable, i.e., $\hat y=-1$. Formally, the set of positive pairs for any sample $x \in \mathbb{Q}$ is:
% To obtain the reliable positive pairs for each anchor $x_i$, function $\mathbb{P}({{x}_{i}})$ has the following format:
\begin{equation}
\mathbb{P}(x_i)=\left\{ \begin{array}{*{35}{l}}
\{{{x}_{j}}\in \mathbb{Q}\setminus\{x_i\}|{\hat{y}_{i}}={\hat{y}_{j}}\} & \text{if}\text{ } x_i\in \mathbb{H},  \\
\{{{x}_{j}}\in \mathbb{Q}\setminus\{x_i\}| \psi(x_i)=\psi(x_j) \} & \text{otherwise}. \\
\end{array} \right.
\label{equation of positive pairs}
\end{equation}

\begin{table*}[ht]
  \centering 
  \resizebox{\textwidth}{!}{
    \begin{tabular}{l|ccc|ccc|ccc}
    \hline
    \multirow{2}{*}{Method}   & \multicolumn{3}{c|}{CIFAR-10} & \multicolumn{3}{c|}{STL-10}           & \multicolumn{2}{c}{SVHN}             \\
    \cline{2-10}
                              & 40         & 250        & 4000 & 40       & 250         & 1000       & 40               & 1000       \\
    \hline
    Fully-Supervised           & \multicolumn{3}{c|}{95.38$\pm$0.05} & \multicolumn{3}{c|}{None}            & \multicolumn{3}{c}{97.87$\pm$0.01}     \\
    \hline
    $\pi$ \text{- Model}~\cite{rasmus2015semi}   & 25.66$\pm$1.76 & 53.76$\pm$1.29 & 86.87$\pm$0.59 & 25.69$\pm$0.85 & 44.87$\pm$1.50 & 67.22$\pm$0.40  & 32.52$\pm$0.95  & 92.84$\pm$0.11 \\
    Pseudo-Label~\cite{lee2013pseudo}           & 25.39$\pm$0.26 & 53.51$\pm$2.20 & 84.92$\pm$0.19 & 25.32$\pm$0.99 & 44.55$\pm$2.43 & 67.36$\pm$0.71 & 35.39$\pm$5.60   & 90.60$\pm$0.32  \\
    Mean Teacher~\cite{tarvainen2017mean}       & 29.91$\pm$1.60 & 62.54$\pm$3.30 & 91.90$\pm$0.21 & 28.28$\pm$1.45 & 43.51$\pm$2.75 & 66.10$\pm$1.37 & 63.91$\pm$3.98  & 96.73$\pm$0.05 \\
    VAT~\cite{miyato2018virtual}                & 25.34$\pm$2.12 & 58.97$\pm$1.79 & 89.49$\pm$0.12 & 25.26$\pm$0.38 & 43.58$\pm$1.97 & 62.05$\pm$1.12 & 25.25$\pm$3.38 & 95.89$\pm$0.20  \\
    MixMatch~\cite{berthelot2019mixmatch}       & 63.81$\pm$6.48 & 86.37$\pm$0.59 & 93.34$\pm$0.26 & 45.07$\pm$0.96 & 65.48$\pm$0.32 & 78.30$\pm$0.68 & 69.40$\pm$8.39  & 96.31$\pm$0.37 \\
    ReMixMatch~\cite{berthelot2019remixmatch}   & 90.12$\pm$1.03 & 93.70$\pm$0.05 & 95.16$\pm$0.01 & 67.88$\pm$6.24 & 87.51$\pm$1.28 & 93.26$\pm$0.14 & 75.96$\pm$9.13  & 94.84$\pm$0.31 \\
    UDA~\cite{xie2020unsupervised}              & 89.38$\pm$3.75 & 94.84$\pm$0.06 & 95.71$\pm$0.07 & 62.58$\pm$8.44 & 90.28$\pm$1.15 & 93.36$\pm$0.17 & 94.88$\pm$4.27  & 98.11$\pm$0.01 \\
    % CCSSL-FixMatch  \cite{yang2022class}        & 70.83 & \textbf{91.60} & 93.96 & 54.50 & \textbf{98.09} & \textbf{98.24}     \\
    MutexMatch~\cite{duan2022mutexmatch}   & 93.22$\pm$2.52 & - & - & - & - & - & 97.19$\pm$0.26  & - \\
    CCSSL~\cite{yang2022class}        & 90.83$\pm$2.78 & 94.86$\pm$0.55 & 95.54$\pm$0.20 & - & - & - & - & - & -\\
    \hline
    FixMatch~\cite{sohn2020fixmatch}           & 92.53$\pm$0.28 & 95.14$\pm$0.05 & 95.79$\pm$0.08  & 64.03$\pm$4.14 & 90.19$\pm$1.04 & 93.75$\pm$0.33 & 96.19$\pm$1.18  & 98.04$\pm$0.03 \\
    % CCSSL-FixMatch*  & 67.78(90)  & 00.00 & 45.13(336) & 97.58(318) & 97.78(484)   \\
    CCL-FixMatch~(ours) & \textbf{94.96}$\pm$0.11$_{\blue{\mbox{\scriptsize +2.43}}}$ & \textbf{95.15}$\pm$0.04$_{\blue{\mbox{\scriptsize +0.01}}}$ & \textbf{95.80}$\pm$0.05$_{\blue{\mbox{\scriptsize +0.01}}}$ & \textbf{70.38}$\pm$3.41$_{\blue{\mbox{\scriptsize +6.35}}}$ & \textbf{91.23}$\pm$0.44$_{\blue{\mbox{\scriptsize +1.04}}}$ & \textbf{94.10}$\pm$0.27$_{\blue{\mbox{\scriptsize +0.35}}}$ & \textbf{97.97}$\pm$0.45$_{\blue{\mbox{\scriptsize +1.78}}}$  & \textbf{98.13}$\pm$0.03$_{\blue{\mbox{\scriptsize +0.09}}}$ \\
    \hline
    \hline
    \end{tabular}
}
% \vspace{-5 pt}
\caption{Top-1 accuracy on CIFAR-10, STL-10, and SVHN datasets. STL-10 dataset does not have label information for unlabeled data, so it does not have fully-supervised results. The \textbf{bold} number indicates the best result. The \blue{blue} values show the improved accuracy of our CCL-FixMatch over FixMatch. We achieve the best performance by simply adding CCL to Fixmatch.}%加了个the
\label{table: main results.}
%\vspace{-10 pt}
\end{table*}

\paragraph{Training objective.}
In the training schema of CCL, we first obtain the pseudo labels and complementary labels for each unlabeled data. Then, we construct reliable negative pairs and positive pairs based on complementary labels. Finally, by optimizing the contrastive loss, the discriminative ability of the model is improved. The contrastive loss $\mathcal{L}_{c}$ takes the following format:
% \begin{small}
\begin{equation}
\begin{split}
         \mathcal{L}_{c}{=}-\sum_{{{x}_{i}}\in \mathbb{Q}}{\frac{1}{|\mathbb{P}({{x}_{i}})|}\sum\limits_{{{x}_{j}}\in \mathbb{P}({{x}_{i}})}{
         \log \frac{{e}\frac{z({{x}_{i}}) \cdot z({{x}_{j}})}{T}}{\sum\limits_{{{x}_{k}}\in \mathbb{P}({{x}_{i}})\cup \mathbb{N}({{x}_{i}})}{{{e}\frac{z({{x}_{i}}) \cdot z({{x}_{k}})}{T}}}}}},
\end{split}
% \begin{split}
%          \mathcal{L}_{c}{=}-\frac{1}{n}\sum_{{{z}_{i}}\in \mathbb{Q}}{\frac{1}{|\mathbb{P}({{z}_{i}})|}\sum\limits_{{{z}_{j}}\in \mathbb{P}({{z}_{i}})}{
%          \log \frac{e^{\left(z_{i}^{\top} z_{j} / T\right)}}{\sum\limits_{{{z}_{k}}\in \mathbb{P}({{z}_{i}})\cup \mathbb{N}({{z}_{i}})}{e^{\left(z_{i}^{\top} z_{k} / T\right)}}}}},
% \end{split}
\label{equation of contrastive loss}
\end{equation}
% \end{small}
where $|\mathbb{P}({{x}_{i}})|$ is the cardinality of the positive pairs set. $z(x_i)$ denotes the feature embedding of $x_i$. $T$ is the temperature coefficient and is set to 0.07 by default.
Besides $\mathcal{L}_c$, we follow~\cite{sohn2020fixmatch, zhang2021flexmatch} and consider two additional loss terms to build the overall training objective. Let $H(\cdot)$ denote the cross-entropy function. We additionally consider a cross-entropy loss on labeled data ${\mathcal{L}}_{x}{=}\frac{1}{B} \sum_{i=1}^{B}{ H\left(y_{i}, \hat{p}_{i}\right)}$ and a loss on unlabeled data ${{\mathcal{L}}_{u}}{=}\frac{1}{\mu B} \sum_{i=1}^{\mu B}{\mathbbm{1}}\left( \max(\hat{p}_i)\ge\tau\right)H\left( \hat{y}_i, f(\mathcal{A}(u_{i})) \right)$.
Thus, the overall loss of our CCL can be written as:
\begin{equation}
       \mathcal{L}=\mathcal{L}_{x}+ \mathcal{L}_{u}+\mathcal{L}_{c}.
\label{equation of total loss}
\end{equation}

\section{Experiments}
In this section, we evaluate the effectiveness of CCL on various SSL datasets. We first describe the main experimental datasets and the training settings in \Cref{dataset}. We then compare CCL with other advanced methods in \Cref{Quantitative Results}. Finally, we present detailed analyses to help understand the advantages of CCL in \Cref{Advantages of CCL}. Code and pretrained models for our CCL will be available soon.

\subsection{Datasets and Training Settings}
\label{dataset}
Following~\cite{ sohn2020fixmatch , yang2022class, zhang2021flexmatch, li2021comatch }, we conduct extensive experiments on several benchmarks, including CIFAR-10~\cite{krizhevsky2009learning}, STL-10~\cite{coates2011analysis}, SVHN~\cite{netzer2011reading}, and CIFAR-100~\cite{krizhevsky2009learning},
each with various amounts of labeled data. Due to page limitation, we put the results on CIFAR-100 in the supplementary. 

\paragraph{Datasets.}
We conduct experiments on CIFAR-10 with 40, 250, and 4000 labeled data, STL-10 with 40, 250, and 1000 labeled data, and SVHN with 40 and 1000 labeled data. CIFAR-10~\cite{krizhevsky2009learning} are all known classes, and the data distribution in each class is balanced.
CIFAR-10 contains 50,000 training images and 10,000 test images and each of the colored images is 32$\times$32 pixels in 10 classes.
STL-10~\cite{coates2011analysis} contains 5000 labeled images and 100,000 unlabeled images. All images are in color and are 96x96 pixels in size. However, the unlabeled images are extracted from a similar but broader distribution of images, which contain unknown classes. 
SVHN~\cite{netzer2011reading} is a relatively simple (i.e., to classify digits) yet unbalanced dataset with 10 classes, which contains a training set, a test set, and 531,131 additional images in the extra set. Note that our experiment also includes the extra set. 
These unlabeled images of SVHN with unbalanced distribution will bring more challenges to the SSL. %in 应该给删了？

\paragraph{Training settings.}
For a fair comparison, we use the same hyper-parameters following the common codebase TorchSSL\footnote{https://github.com/TorchSSL/TorchSSL}~\cite{zhang2021flexmatch}. We use Wide ResNet-28-2 \cite{zagoruyko2016wide} for CIFAR-10, STL-10, and SVHN. In the testing phase, we perform all algorithms using an exponential moving average with the momentum of $0.999$. In a mini-batch, 64 labeled data and $\mu\times 64$ unlabeled data are randomly sampled, where the unlabeled data ratio $\mu$ is set to $7$. The optimizer for all experiments is standard stochastic gradient descent (SGD) with the momentum of $0.9$. For all datasets, we use cosine learning rate decay schedule \cite{loshchilov2016sgdr} as $\eta=\eta_{0} \cos \left(\frac{7 \pi t}{16 N}\right)$, where $\eta_{0}=0.03$ is the initial learning rate, $t$ is the current training step, and $N$ is the total training step that is set to $N=2^{20}$ for all datasets. The default confidence threshold $\tau$ is set to 0.95. The default number of classes in complementary labels $k$ is set to 7. In all of our experiments, the weak augmentation $\mathcal{\alpha}(\cdot)$ is a random flip-and-shift augmentation strategy, and the strong augmentation $\mathcal{A}(\cdot)$ is RandAugment~\cite{cubuk2020randaugment}. Following \cite{zhang2021flexmatch}, we run each task three times using random seed 0, 1, 2 for all experiments and report the best top-1 accuracy of all checkpoints. Detailed hyper-parameters are introduced in the supplementary.

\subsection{Comparisons on Standard Benchmarks}
\label{Quantitative Results}
We compare our method with some advanced SSL methods, including $\pi$ \text{- Model} \cite{rasmus2015semi}, Pseudo-Label \cite{lee2013pseudo}, Mean Teacher \cite{tarvainen2017mean}, VAT \cite{miyato2018virtual}, MixMatch \cite{berthelot2019mixmatch}, ReMixMatch \cite{berthelot2019remixmatch}, UDA \cite{xie2020unsupervised}, MutexMatch \cite{duan2022mutexmatch},  CCSSL~\cite{yang2022class} and FixMatch \cite{sohn2020fixmatch}. Meanwhile, we also report the results of full-supervised learning with all data labeled on CIFAR-10~\cite{krizhevsky2009learning} and SVHN~\cite{netzer2011reading}.  
%Since STL-10 doesn’t have the label corresponding to its 100,000 unlabeled data, it’s unavailable to apply full-supervised learning on it. 
What's more, we also provide the detailed precision, recall, F1, and AUC results in the supplementary. % Deng：这句还要根据实验结果斟酌下要不要

\begin{table*}[!ht]
\centering 
    \begin{tabular}{ccc|c}
    \hline
            \thead{Constructing Pairs in $\mathbb{H}$}      & \thead{Constructing Pairs in $\mathbb{L}$}           & \thead{Constructing Pairs with Complementary \\Labels Based on \Cref{equation of negative pairs} and \Cref{equation of positive pairs}}                   & \thead{CIFAR-10 w/ 40}   \\
    \hline                          
       $\times$         & $\times$              & $\times$                 & 92.53 (baseline)       \\
       \checkmark       & $\times$              & $\times$                 & 93.68$_{\blue{\mbox{\scriptsize +1.15}}}$   \\
       \checkmark       & \checkmark            & $\times$                 & 93.03$_{\blue{\mbox{\scriptsize +0.50}}}$ \\
        \checkmark       & \checkmark            & \checkmark               & \textbf{94.96}$_{\blue{\mbox{\scriptsize +2.43}}}$     \\
    \hline
    \end{tabular}
\caption{Comparison of using different strategies for constructing pairs in contrastive learning on CIFAR-10 with 40 labeled data. The \blue{blue} values show the improved accuracy of our CCL-FixMatch over FixMatch (baseline). Results show that the strategy of using complementary labels to construct reliable negative pairs for contrastive learning can significantly improve model performance.}
%Using complementary labels to construct pairs in contrastive learning can effectively use the information of unreliable unlabeled data.}
\label{module ablation}
% \vspace{-10 pt}
\end{table*}

In \Cref{table: main results.}, we find that CCL is helpful for FixMatch for all benchmarks and CCL achieves better performance when the task contains more noise (i.e., fewer labels). For example, under the setting of CIFAR-10 with only 4 labeled data per class, CCL achieves $+2.43\%$ on FixMatch. On unbalanced dataset SVHN, CCL-FixMatch also improves $1.78$\% and $0.09$\% performance over FixMatch for 40 and 1000 labels. STL-10 dataset is a more complicated and more realistic task since the existence of unknown classes in the unlabeled data. However, in such a challenging case, CCL-FixMatch achieves $+6.35$\%, $+1.04$\% and $+0.35$\% accuracy improvement on FixMatch of STL-10 with 40, 250 and 1000 labeled data, respectively. 
The decrease in performance gain can be explained as the increasing discriminative ability of the model with the help of more labeled training data.
%The decrease in performance gain can be explained as the discriminative ability that the model increases with more labeled training data
And a well-performing model can also be obtained in less noisy settings without the use of contrastive learning. These improvements also demonstrate the capabilities and potential of CCL for real-world applications.%incresing 应该是do还是doing

\subsection{Advantages of CCL over Existing Methods}
\label{Advantages of CCL}
\begin{figure}[!t]
\centering
\includegraphics[width=1\linewidth]{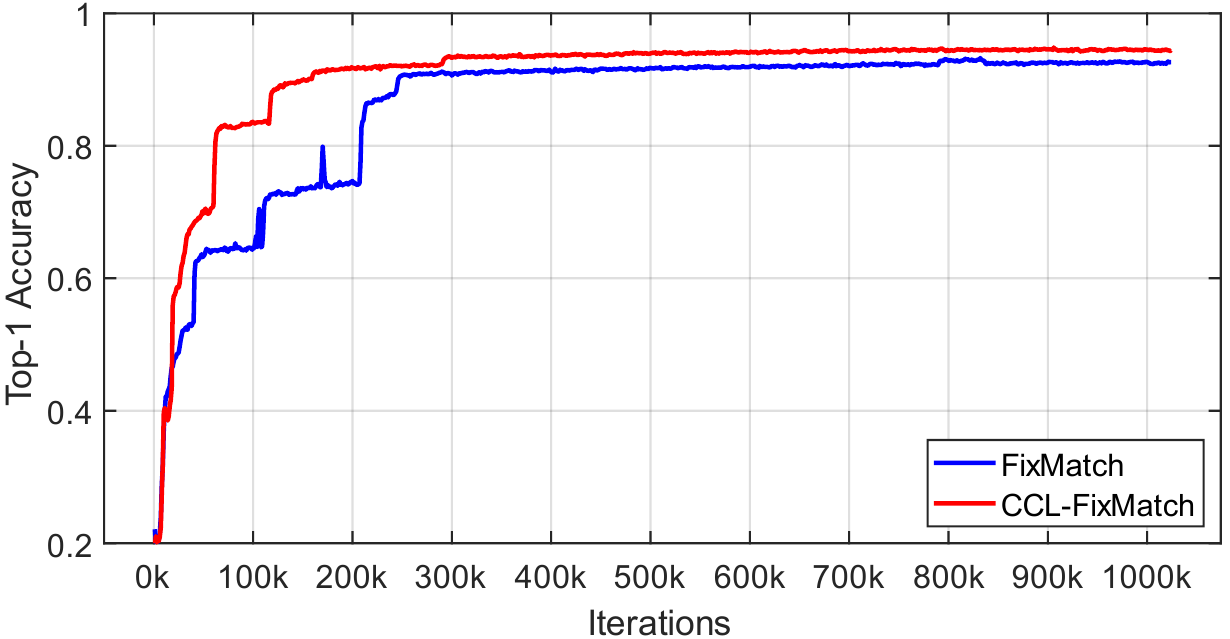} 
\caption{Comparison of the accuracy of FixMatch and CCL-FixMatch on CIFAR-10 with 40 labeled data. CCL achieves better performance and the model is faster in global convergence.} 
\label{figure: acc_figure}
\vspace{-10 pt}
\end{figure}

\paragraph{Faster convergence with CCL.}
In \Cref{figure: acc_figure}, We study the training convergence speed between FixMatch and CCL-FixMatch on CIFAR-10 with 40 labeled data. CCL-FixMatch achieves better performance with fewer training iterations and shows its superior convergence speed. By using contrastive learning with complementary labels, the model can place the positive pairs as close as possible and the negative pairs as far away as possible in the embedding space, which helps to learn a good feature distribution among instances and improves the discriminative ability of the model in the early training stage. Through our reliable pair construction strategy, the model can better utilize the information of unreliable unlabeled data to guide the learning process and get faster convergence. Therefore, with only 300k iterations, CCL-FixMatch has already surpassed the final results of FixMatch.

\begin{figure}[!t]
  \centering
  \vspace{4.7mm}
  \begin{subfigure}{0.49\linewidth}
    \includegraphics[width=1\linewidth]{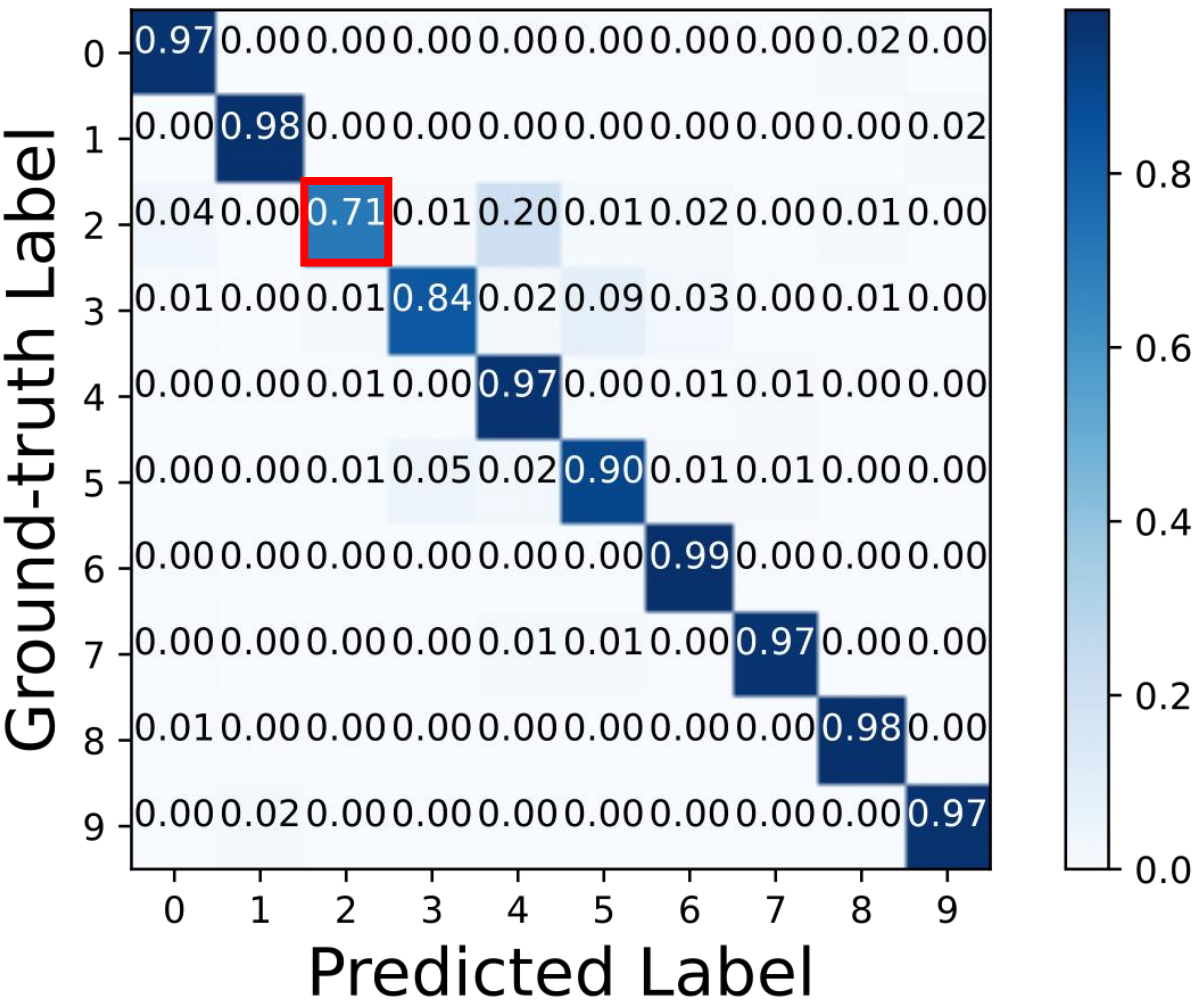}
    % \caption{Confusion matrix of FixMatch.}
    \label{subfig:confusion_matrix of fixmatch}
  \end{subfigure}
  \hfill
  \begin{subfigure}{0.49\linewidth}
    \includegraphics[width=1\linewidth]{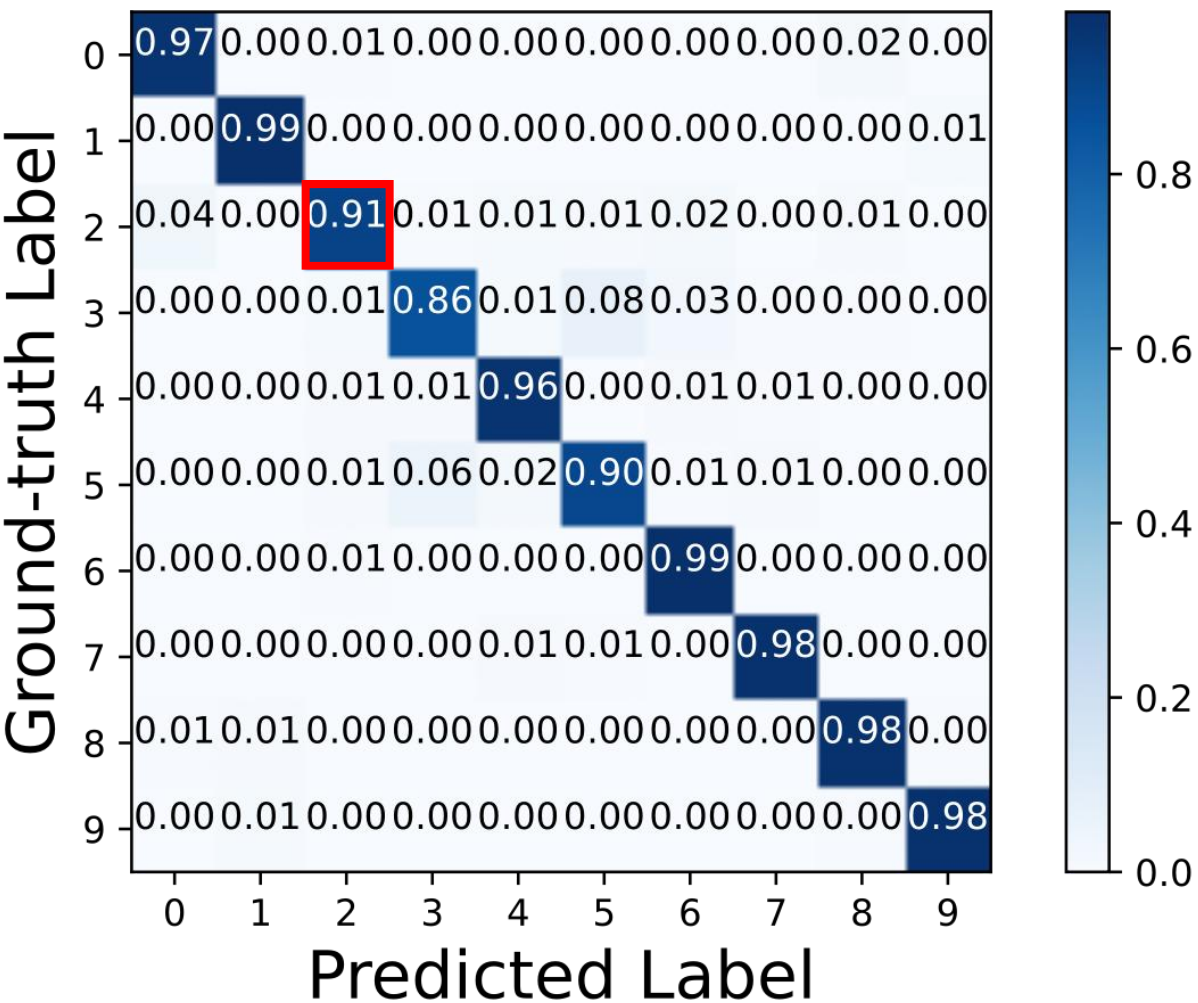}
    % \caption{Confusion matrix of CCL-FixMatch.}
    \label{subfig:confusion_matrix of ccl-fixmatch}
  \end{subfigure}
  \caption{Confusion matrices obtained based on the best checkpoints of FixMatch (left) and CCL-FixMatch (right) on CIFAR-10 with 40 labeled data. The \red{red box} indicates the hard class on which FixMatch yields a very low accuracy. By contrast, our CCL model tends to learn a more balanced distribution across different classes.}
  \label{fig:confusion_matrix}
\end{figure}

\paragraph{Superior performance on hard classes.}
As shown in \Cref{fig:confusion_matrix}~(left), the confusion matrix results show that FixMatch is difficult to learn on hard classes. For instance, there are more than 20\% examples that belong to class 2 (bird) that are predicted wrongly as class 4 (deer). And CCL obtains a more unbiased confusion matrix as shown in \Cref{fig:confusion_matrix}~(right). CCL can improve the inter-class separation degree, which promotes the learning of hard classes by the model, improve the quality of pseudo labels, and help to improve the generalization performance.%加个that?

\begin{table*}[!t]
  \centering 
  \resizebox{\textwidth}{!}{
    \begin{tabular}{l|ccc|ccc}
    \hline
    \multirow{2}{*}{Method}                 & \multicolumn{3}{c|}{CIFAR-10}               & \multicolumn{3}{c}{STL-10}     \\
    \cline{2-7}
                             & 10          & 20          & 40               & 10               & 20               & 40              \\
    \hline
    FixMatch~\cite{sohn2020fixmatch}     & 46.47$\pm$20.50          & 85.81$\pm$7.24           & 92.53$\pm$0.28          & 34.43$\pm$2.37          & 46.89$\pm$2.96    & 64.03$\pm$4.14    \\
    CCL-FixMatch (ours)                  & \textbf{49.80}$\pm$20.37$_{\blue{\mbox{\scriptsize +3.33}}}$ & \textbf{88.74}$\pm$1.27$_{\blue{\mbox{\scriptsize +2.93}}}$ & \textbf{94.96}$\pm$0.11$_{\blue{\mbox{\scriptsize +2.43}}}$ & \textbf{40.51}$\pm$2.13$_{\blue{\mbox{\scriptsize +6.08}}}$  & \textbf{50.79}$\pm$2.09$_{\blue{\mbox{\scriptsize +3.90}}}$   & \textbf{70.38}$\pm$3.41$_{\blue{\mbox{\scriptsize +6.35}}}$    \\
    \hline
    FlexMatch~\cite{zhang2021flexmatch}   & 77.85$\pm$17.39          & 93.20$\pm$2.06     & 95.03$\pm$0.06             & 43.99$\pm$3.82             & 57.63$\pm$1.89   & 70.85$\pm$4.16     \\
    CCL-FlexMatch (ours)                  & \textbf{82.88}$\pm$16.82$_{\blue{\mbox{\scriptsize +5.03}}}$ & \textbf{94.60}$\pm$0.48$_{\blue{\mbox{\scriptsize +1.40}}}$     & \textbf{95.04}$\pm$0.06$_{\blue{\mbox{\scriptsize +0.01}}}$    & \textbf{44.39}$\pm$3.74$_{\blue{\mbox{\scriptsize +0.40}}}$   & \textbf{57.98}$\pm$1.52$_{\blue{\mbox{\scriptsize +0.35}}}$    & \textbf{76.09}$\pm$2.05$_{\blue{\mbox{\scriptsize +5.24}}}$    \\
    \hline 
    \end{tabular}
}
\caption{Comparison of the performance of different methods on CIFAR-10 and STL-10 only with 10, 20, and 40 labeled data, respectively. The \blue{blue} values show the improved accuracy of our CCL-FixMatch over FixMatch. CCL-FixMatch/CCL-FlexMatch significantly improves the accuracy and model stability~(i.e. smaller standard deviation) of FixMatch/FlexMatch under the label-scare settings. }
\label{table: fewer label}
\end{table*}

\section{Ablations and Further Discussions}
In \Cref{ablation:Different Strategies of Pairs Construction}, We first analyze the influence of different strategies of pairs construction in contrastive learning. Then, in \Cref{ref:Model Performance under Label-Scarce Setting}, we further explore the applicability of our method under the label-scare settings. Finally, we investigate the effect of the number of classes to construct complementary labels $k$ in \Cref{alation:k}.

\subsection{Different Strategies of Pairs Construction}
\label{ablation:Different Strategies of Pairs Construction}
In \Cref{module ablation}, we investigate the effect of contrastive learning with different strategies for pair construction on CIFAR-10 with 40 labeled data. We find that, based on the FixMatch, simply using the pseudo labels of each instance in the high-confidence set $\mathbb{H}$ as supervision information, and introducing supervised contrastive learning~\cite{khosla2020supervised} to construct positive and negative pairs, can obtain the accuracy of 93.68\%, improving the accuracy of FixMatch by 1.15\%.
However, if we further use the unreliable low-confidence set $\mathbb{L}$ to construct pairs, the performance is improved compared to the baseline but is slightly degraded compared to the strategy of constructing pairs using only $\mathbb{H}$.
This suggests that contrastive learning can be effectively applied to SSL when the data is relatively reliable. Therefore, to effectively utilize unlabeled data and reduce noise, we use complementary labels to construct reliable pairs for unreliable low-confidence data, and then use contrastive learning to improve the discriminative ability of the model. With this strategy, the final model improves the performance of SSL and achieves an accuracy of $94.96\%$.

\begin{figure}[!t]
\centering
\includegraphics[width=1\linewidth]{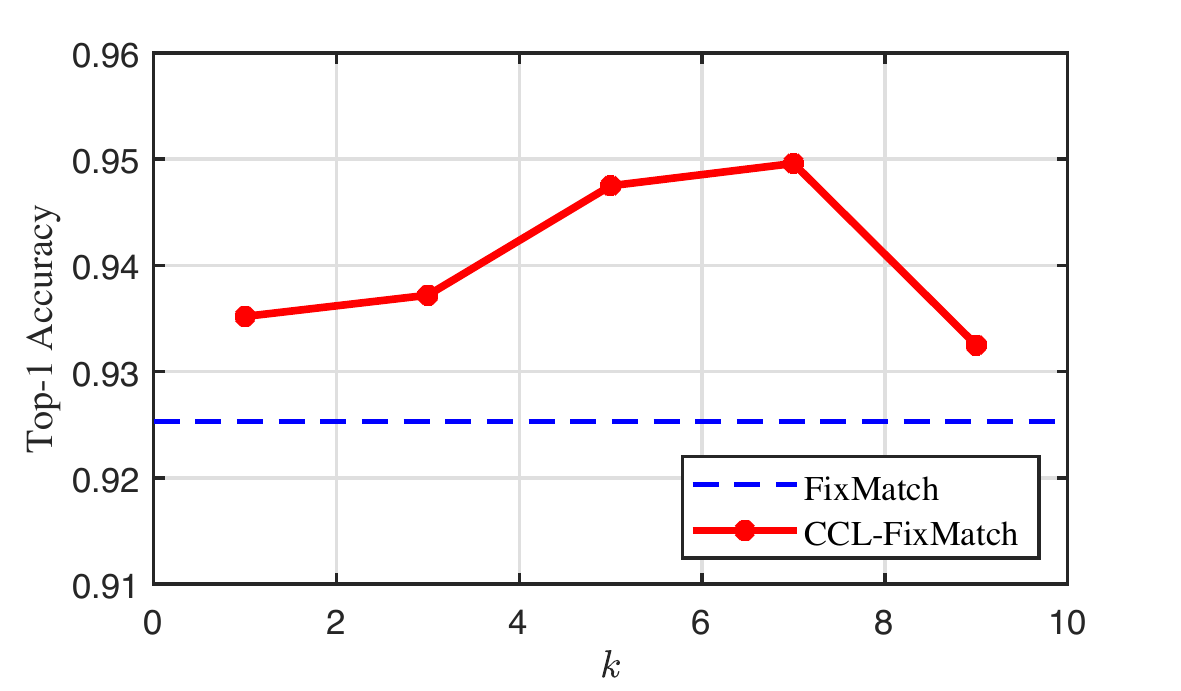} 
\caption{Impact of the number of classes to construct complementary labels, i.e., $k$, on CIFAR-10 with 40 labeled data based on CCL-FixMatch.
% Experiment on the effect of choosing the different number of classes to construct complementary labels $k$ on CIFAR-10 with 40 labeled data in CCL-FixMatch. 
% A moderate value of $k$ (e.g. $k=7$) brings the improvement in performance.
No matter how we change the value of $k$, we always obtain better results than the baseline model, thanks to the additional information provided by low-confidence samples.
In practice, we obtain the best result with $k=7$.
} 
\label{figure: different k}
\vspace{-10 pt}
\end{figure}

\subsection{Comparisons under the Label-Scarce Settings}
\label{ref:Model Performance under Label-Scarce Setting}
In \Cref{table: fewer label}, the results demonstrate that CCL can significantly improve the performance of FixMatch~\cite{sohn2020fixmatch} and FlexMatch~\cite{zhang2021flexmatch} under the label-scarce setting. 
CCL-FixMatch consistently outperforms FixMatch under the settings with extremely limited labeled data. For example, on CIFAR-10 with 10, 20, and 40 labeled data, the accuracy is improved by 3.33\%, 2.93\%, and 2.43\%, respectively. On STL-10 with 10, 20, and 40 labeled data, CCL-FixMatch outperforms Fixmatch by 6.08\%, 3.90\%, and 6.35\%, respectively. Compared to FixMatch which uses fixed predefined thresholds, FlexMatch uses flexible class-specific thresholds. And CCL-FlexMatch can further improve the performance of FlexMatch. For example, on CIFAR-10 with 10, 20 and 40 labeled data, the accuracy improves from 77.85\% to 82.88\%, from 93.20\% to 94.60\% and from 95.03\% to 95.04\%, respectively. On STL-10 with 10, 20, and 40 labeled data, CCL-FlexMatch improves the performance by 0.40\%, 0.35\%, and 5.24\% respectively compared to the accuracy of FlexMatch. In brief, CCL can utilize contrastive learning with complementary labels to effectively alleviate the accumulation of possible noises in the training process.%确定一下结论是否正确

\subsection{Effect of $k$ on Model Performance}
\label{alation:k}
In \Cref{figure: different k}, we also study different values of the number of classes to construct complementary labels $k$ on CIFAR-10 with 40 labeled data. We find that the best practice value for $k$ is 7 and that too high or too low $k$ can degrade model performance. The reason for this phenomenon is that different values of $k$ affect the number of negative pairs constructed. Specifically, stricter negative pair construction strategies (i.e. smaller $k$) can reduce noise as much as possible, but a lot of unlabeled data information is reduced to guide model training. On the contrary, too loose negative pair construction strategies (i.e. larger $k$) will increase the number of construction of incorrect negative pairs, which will introduce too much noise in contrastive learning.

\section{Conclusion}
In this paper, we study how to better leverage unreliable low-confidence unlabeled data but reduce possible noises. To address this issue, we propose a Contrastive Complementary Labeling (CCL) method, which constructs reliable negative pairs based on the complementary labels of low-confidence data. In this way, CCL provides discriminative information for subsequent contrastive learning without introducing possible noises. Extensive experimental results on multiple benchmark datasets show that CCL greatly improves the performance of existing advanced methods, especially under the label-scarce settings.

%%%%%%%%% REFERENCES
{\small
\bibliographystyle{ieee_fullname}
\bibliography{egbib}
}

\end{document}